# RESOLVING OPEN-TEXTURED RULES WITH TEMPLATED INTERPRETIVE ARGUMENTS

JOHN LICATO
*Advancing Machine and Human Reasoning (AMHR) Lab*
*Department of Computer Science and Engineering*
*University of South Florida*
licato@usf.edu

LOGAN FIELDS
*Advancing Machine and Human Reasoning (AMHR) Lab*
*Department of Computer Science and Engineering*
*University of South Florida*
ldfields@usf.edu

ZAID MARJI
*Advancing Machine and Human Reasoning (AMHR) Lab*
*Department of Computer Science and Engineering*
*University of South Florida*
zaidm@usf.edu

**Abstract**

Open-textured terms in written rules are typically settled through interpretive argumentation. Ongoing work has attempted to catalogue the schemes used in such interpretive argumentation. But how can the use of these schemes affect the way in which people actually use and reason over the proper interpretations of open-textured terms? Using the interpretive argument-eliciting game Aporia as our framework, we carried out an empirical study to answer this question. Differing from previous work, we did not allow participants to argue for interpretations arbitrarily, but to only use arguments that fit with a given set of interpretive argument templates. Finally, we analyze the results captured by this new dataset, specifically focusing on practical implications for the development of interpretation-capable artificial reasoners.

## 1 Introduction

When people delegate decision-making authority to others, that delegation is often accompanied with guidelines in the form of *open-textured rules*—i.e., rules containing open- textured terms, such that the exact interpretations of those terms can be delegated to the discretion of boots-on-the-ground decision-makers. In this way, the intent of the rule-writer can be formalized such that it provides guidance but is not overly constraining, to avoid placing unnecessary constraints on the adaptability of the rule-interpreters. For example, a social media company may instruct its employees to "not allow users to use our website in a manner intended to artificially amplify or suppress information." Should the situation arise, the employees with the responsibility of executing this directive must determine the proper interpretation of the open-textured term 'artificially amplify'. Suppose a university decides to post an announcement and has it re-tweeted through the Twitter accounts of its multiple academic departments, colleges, alumni groups, etc. Would this action constitute artificial amplification in the sense meant by the rule? Furthermore, suppose that the employee is not human at all, but an automated Twitter moderation-bot. How should an answer to the preceding question be searched for and justified, in such a way that is satisfactory to the rule-writers? Such a situation may not have been anticipated by the rule-writers at the time of rule writing, and so the attempt to state the rule in a form general enough to sufficiently cover unanticipated scenarios can result in open-textured language.

For the most part, the use of open-textured rules to guide human behavior works because of the assumption that when the rule-interpreters interpret the rules, it will likely be done in a manner that is sufficiently similar to the interpretation that would be used by the rule-writer under similar circumstances. But this assumption cannot be made when the rule-writers are human beings and the rule-interpreters are artificially intelligent reasoners. It is therefore crucial that serious research be carried out into developing automated interpretive reasoning techniques that can give us some guarantee of human likeness.

One appealing approach is to require interpretations of open-textured rules to be justified with a graph of human-like interpretive arguments, properly organized according to some norms of argument strength and combination (possible frameworks for which are well-known to the AI in argumentation community). If reasonable effort is placed into generating the strongest possible interpretive arguments for and against a possible interpretation (given constraints of time, computational power, and available information), comparing them to each other, and combining them, the resulting interpretation would have some claim to being the most rational, or at least satisficing. It is our belief that such an approach to automated interpretive reasoning is significantly preferable to current approaches in AI; specifically, the so-called "explainable AI" paradigm which seeks to explain how an AI black box reaches its conclusion, rather than seeking to justify that conclusion with normative argumentation. This is the approach argued by the MDIA view, which states that rule-following AI should act in accordance with the interpretation best supported by *minimally defeasible interpretive arguments*—i.e., those interpretive arguments which are such that there are a minimal number and quality of interpretive counterarguments that can be levied against them (Licato, 2021).

Taking some variant of the MDIA position as our base assumption,[1] we seek to advance progress towards interpretation-capable AI: First, we describe Aporia, a new tool for eliciting and comparing naturalistic interpretive arguments. Second, we present a new dataset, collected using Aporia, with the caveat that all interpretive arguments used are required to utilize a fixed set of interpretive argument templates. Finally, we analyze the results captured by this new dataset, specifically focusing on practical implications for the development of interpretation-capable artificial reasoners.

## 2 Experimental Setup

Aporia (Marji et al., 2021) is a gamified framework to elicit naturalistic interpretive arguments and comparisons of interpretive arguments. It begins with an open-textured rule (e.g., “No vehicles are allowed in the park”) and an intentionally ambiguous scenario (e.g, “A non-working antique car, meant for an artistic exhibit, is pushed into the park”). Three players—Player 1, Player 2, and the judge—are then asked to determine whether the rule applies. Player 1 is asked to first decide on a position: either that the action described in the scenario is a violation of the rule, or that the rule does not apply. Player 1 must then argue their position. Next, Player 2 is given time to counter Player 1’s arguments. Finally, the judge is asked to determine whether Player 1 was more convincing, or whether Player 2 cast sufficient doubt onto Player 1’s argument that it cannot be accepted, and to provide their rationale for their decision.

[1] We will not mount a full defense of the MDIA position, instead relying on the defense in [1].

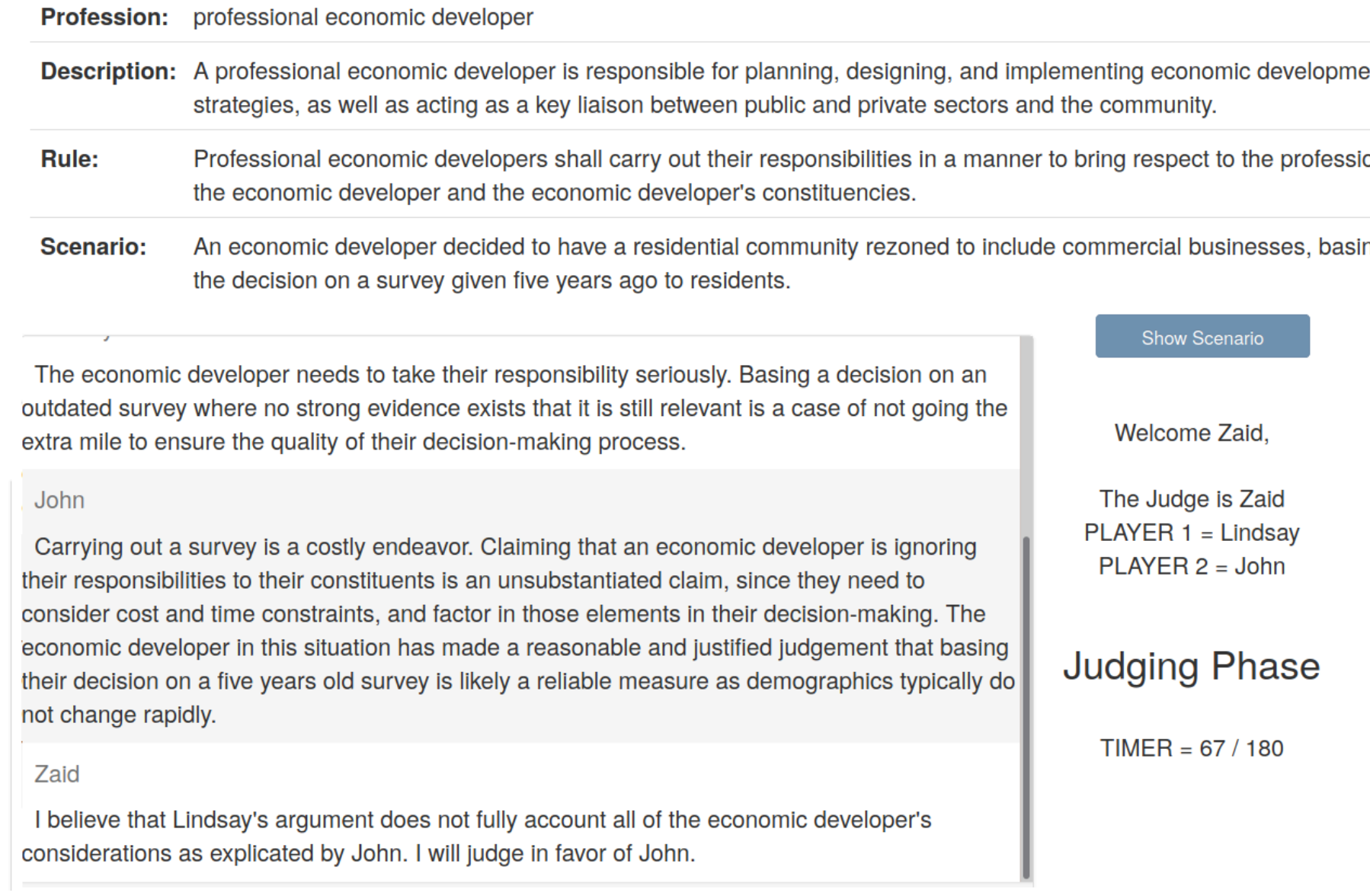

**Figure I.** Example of Aporia Gameplay

Previously, Aporia has been used to collect a dataset of played games, seeded with rules and scenarios originally collected from codes of ethics of various professional societies (Licato et al., 2019). However, this work only included one judgment per argument pair, and thus does not allow us to rigorously study the range of judgments used to compare interpretive reasoning. Secondly, it did not enforce the use of interpretive argument templates, thus making annotation and comparison of interpretive argument types difficult.

We therefore adopted a typology of interpretive argumentation schemes, primarily adapted from Walton et al. (2021) and simplified to be: (1) more accessible to laypeople (i.e., non-experts in legal reasoning), (2) less fine-grained, and (3) de-emphasize arguments rooted in specific legal practices, so as to be more broadly applicable to general interpretive reasoning of rules. The resulting argument types we employed are:

1. **Argument from Authoritative Source:** An authoritative source (an official definition, authoritative document, or authority figure / expert) defines a term a certain way, which then requires a certain interpretation. **Critical questions:** Does the authoritative source actually say what is claimed? Is the authority's area of expertise or jurisdiction relevant to the rule? Is there another authority that has an alternative interpretation?
2. **Argument from Higher Reason or Principle:** There is a higher principle of reasoning, ethics, or law that requires us to interpret the rule a certain way. **Critical questions:** Is the claimed higher principle actually used in practice? Does it actually govern our interpretations? Is there another example where the claimed higher principle is clearly violated?

3. **Argument from Practicality / Consequence:** It is more effective, efficient, or otherwise practical to interpret the rule in a certain way when considering certain consequences of alternate interpretations.
   **Critical questions:** Would this interpretation lead to a bending of the rule that would render the rule useless? Is the claimed consequence actually going to happen with any certainty, or does the new level of certainty justify the new interpretation?
4. **Argument from History:** The historical interpretation of a term is, or has evolved in, a specific way that constrains how we can interpret it today.
   **Critical questions:** Is the claimed historical interpretation actually accurate? Is there a reason to believe that the historical interpretation is outdated and no longer in use?
5. **Argument from Public Understanding:** The public, commonsense, reasonable, or de- fault understanding of a term is such that it requires a certain interpretation.
   **Critical questions:** Is the public understanding of the term claimed here actually correct, or is the arguer mistaken? Are there other understandings of the term that lead to different interpretations, but which are just as public, commonsense, reasonable, or default?
6. **Argument by Analogy / Similarity:** If the action being considered (A1) is similar to another previous action that the rule does or does not apply to (A2), then we can use that to determine whether the rule applies to A1.
   **Critical questions:** What kind of similarity does A1 and A2 have? Is it actually relevant to the rule? Are there important differences or dissimilarities between A1 and A2? Is there another similar case (A3) that is also similar to A1, except that the rule applies differently to A2 than it did to A3?
7. **Argument from Word Choice / Exclusion:** If the rule is phrased in a very specific way that leaves out certain words or terms, then there is a reason for it. And that reason requires us to interpret the rule in a certain way.
   **Critical questions:** Is the omission explainable by some other reason? E.g., is it possible that the rule writers did not think it was necessary to include because it was to be assumed, or inferred?

We then created six rule-scenario pairs. All rules were taken directly from Twitter's rules of conduct in March 2022 (prior to current substantial management changes in that organization)[2] and scenarios were designed to intentionally exploit the open-texturedness of the rules:

1. **Rule:** You may not promote violence against, threaten, or harass other people on the basis of race, ethnicity, national origin, caste, sexual orientation, gender, gender identity, religious affiliation, age, disability, or serious disease.
   **Scenario:** You retweet a position held by violent extremists, without commentary, and frequently retweet similar opinions from other people. Your account has a large number of followers.

[2] https://help.twitter.com/en/rules-and-policies/twitter-rules

2. **Rule:** You may not promote or encourage suicide or self-harm.
   **Scenario:** You never seriously encourage someone to commit suicide or self-harm, and would go out of your way to prevent someone from doing those things, but you publicly joke and laugh about suicide-related jokes.
3. **Rule:** You may not engage in the targeted harassment of someone, or incite other people to do so. This includes wishing or hoping that someone experiences physical harm.
   **Scenario:** A group of reddit users work together to solve a crime. They put together evidence and conclude that person X is guilty, posting images, personal information, and other details publicly to support their investigation and claims. However, X is later proven to be innocent. Did the group participants harass X?
4. **Rule:** You may not use Twitter's services in a manner intended to artificially amplify or suppress information.
   **Scenario:** You work for the University's social media team, and when a faculty member earns a new grant, you post it on the social media pages of the department, college, university, alumni group, faculty group, and ask friends and colleagues to like and re-share the posts.
5. **Rule:** You may not impersonate individuals, groups, or organizations to mislead, confuse, or deceive others, nor use a fake identity in a manner that disrupts the experience of others on Twitter.
   **Scenario:** You create an account called "Donald Drumpf" meant to mock things the former president does, by simply re-posting things he actually said as tweets, without additional context.
6. **Rule:** You may not threaten violence against an individual or a group of people. We also prohibit the glorification of violence.
   **Scenario:** You frequently post positive memes, thoughts, etc. honoring the "greatest generation," who are named this because they fought in, and won, WWII.

Thirty undergraduate students at the University of South Florida were assigned as follows: thirty games were created, each one employing one of the rule-scenario pairs above, such that each rule-scenario pair was assigned to exactly five games. Each student was first assigned to one game, asked to choose a position to defend, and create two arguments in support of their position. Each argument was required to utilize one of the templates, instructions for which were provided to them in randomized order. The students were then shuffled and assigned to another game, and asked to address the first student's arguments—not to simply devise arguments for the opposite position, but to focus on undercutting or undermining Player 1's argument (thus differing slightly from Marji et al. (2021), which instead encouraged general counterarguments). Students were then re-assigned again, and asked to judge which arguments were more convincing for a third game, to explain why, and rate their confidence in their decision on a scale from 1 to 5 (5 being the highest confidence level).

This last step was repeated three more times, so that each game had four unique judgments (in order to ensure each judgment was made independently, each judge was unable to see the previous judges' decisions). Every player was assigned such that for every one of their six rounds (initial argument, counterargument, and four judging rounds), they were assigned to a game with a different rule-scenario pair.

**Table I.** Evaluations of whether argument schemes were correctly applied.

| Template name | Classified as | | |
|---|---|---|---|
| | **1** | **2** | **3** |
| Public Understanding | 0 | 3 | 11 |
| Word choice or exclusion | 1 | 6 | 7 |
| Practicality or consequence | 2 | 2 | 9 |
| Authoritative source | 0 | 3 | 1 |
| Analogy or Similarity | 0 | 1 | 3 |
| History | 0 | 0 | 3 |
| Higher reason or principle | 1 | 0 | 4 |
| **Total** | **4** | **15** | **38** |

In order to determine how well argument schemes were correctly applied, three Advancing Machine and Human Reasoning Lab researchers (the present paper's authors: one faculty member and two PhD students), each with several years of experience in working with interpretive argumentation, independently assessed the arguments provided by participants. They were instructed to categorize each argument into one of the following categories: (1) the annotator could not understand the structure of the argument and figure out how it fit into its supposed argument scheme after reading it at least three times; (2) the annotator identified the structure of the argument being made, but it was of a different scheme than the one it was claimed to be; and (3) the structure of the argument being made is closest to the argument scheme it was claimed to be. The resulting evaluations are listed in Table I. Note that due to technical errors, three arguments were excluded from the analysis, thus the bottom row of Table I totals to 57 rather than 60 (these excluded arguments are also excluded from all statistics we report for the remainder of this paper).

All arguments for all stages were randomly assigned to the annotators so that each argument was categorized by two annotators independently. If the annotators for any argument disagreed on the proper category, then the third annotator would independently categorize it. In such cases, the category chosen by two out of the three annotators would be counted as the correct one. In no cases did the three annotators select three different categories.

## 3 Analysis

We selected our rule-scenario pairs in order to encourage alternate interpretations, so that we could focus on the argumentation used to justify interpretations. Since Player 1 in each game was allowed to choose which position they would defend (*violation* or *non-violation*), the relative number of times players chose each position for each rule-scenario pair is a way of estimating how well they invite competing interpretations. According to this measure, we were partially successful: rule-scenario pairs 3, 4, and 6 were as close to perfectly balanced as possible (3 players chose *violation*, and 2 chose *non-violation*, or vice-versa). All players chose *non-violation* for rule-scenario pairs 1 and 5, and all chose *violation* for pair 2.

**Table II.** Frequency of argument types used

| Template name (**L**inguistic, **S**ystemic, or **T**eleological-Evaluative) | Used as 1st arg | Used as 2nd arg | Total |
| --- | --- | --- | --- |
| Public Understanding (L) | 10 | 4 | 14 |
| Word choice or exclusion (L) | 6 | 8 | 14 |
| Practicality or consequence (S) | 7 | 6 | 15 |
| Authoritative source (S) | 3 | 1 | 4 |
| Analogy or Similarity (S) | 1 | 3 | 4 |
| History (S) | 0 | 2 | 2 |
| Higher reason or principle (TE) | 1 | 4 | 5 |

MacCormick and Summers (1991) suggested that interpretive arguments be considered and deployed with linguistic arguments coming first, systemic arguments next, and teleological-evaluative arguments last. Table II categorizes and sorts our argument schemes in this order, in order to easily see whether arguers naturally selected arguments in this order. Indeed, linguistic arguments were most commonly chosen as the first argument type, with systemic arguments the second most common choice for first arguments. Given that our arguers were not formally trained in legal reasoning or formal argumentation, it is interesting to see MacCormick and Summers' ordering reflected (albeit weakly) in the arguers' preferences.

We further set out to examine whether the evaluations of argument quality, as made by our participants acting in the role of *judge*, were consistent. We consider a simple majority vote of judges assigned to a game as deciding whether Player 1 or 2 was the winner of that game. Across all 30 games, we see that Player 1 was the winner 53.6% of the time, Player 2 won 32.1% of the time, and there was a tie 14.3% of the time, suggesting that although there is a preference for Player 1, the advantage that Player 1 gets from being able to go first is not non-existent.

A similar pattern manifests when we aggregate the individual judge decisions together, regardless of game: 58.7% and 41.3% of judge decisions were for Player 1 and 2, respectively. Furthermore, when judges selected Player 1 as the winner, they had an average confidence of 4.25 (out of 5). However, when selecting Player 2, their average confidence dropped to 3.76. This may reflect a hesitation to go against the first arguments they read (those of Player 1), consistent with known anchoring and ordering effects in argumentation. However, our study design did not allow for us to vary the order in which players' arguments were read, since Player 2's arguments were always responses to those of Player 1.

In order to determine whether shallow heuristics might have been employed to determine argument quality, we calculated the word length of arguments and used Spearman correlation to compare them to the proportion of judges that voted for them. This effect was significant ($r=0.534$, $p < 0.005$), suggesting that arguments that had more words were considered more persuasive.

Furthermore, we calculated the correlation between the average confidence of judges in their decisions (recall that they self-rated their confidence from 1 to 5) and the level of agreement between judges on that game (defined as the proportion of judges who voted for the majority vote, or zero if there was a tie). The correlation was small ($r=0.347$, $p<0.08$), weakly suggesting that when a game's two players produced competing arguments that were difficult to decide between (in the sense that a sample of judges will disagree as to which is more convincing), judges were able to anticipate this controversiality and it lowered their confidence in their ratings.

# 4 Conclusion and Future Work

It is not fully known how controlled argumentation dialogue environments, such as *Aporia*, affect the types of argumentation used. We expect that, at least among non-experts, unrestricted argumentation environments will result in arguments that are more difficult to classify as one of the known interpretive argument types. The work presented in this paper offers a data point that can be used as a baseline against which to compare future studies.

In particular, it would be interesting to see how experience with using interpretive argumentation changes the way in which they are used. Table I summarized our evaluations on whether argument schemes were applied correctly. In our study, arguments were rated as category 3 (meaning they were applied correctly) 66.1% of the time, a value we attribute to the short amount of time that arguers were given to familiarize themselves with the argument schemes (roughly 10 minutes).

An under-studied aspect of interpretive reasoning is how experts and non-experts compare and evaluate interpretive arguments. Our present work offers some empirical insight into this---e.g., we showed that argument length correlates with judges' perceptions of argument quality, and the confidence of judges also negatively correlates with how controversial they think their decision will be. But are there deeper relationships between the rationales used by judges in evaluating interpretive arguments and their ultimate decisions? Since the present study required judges to provide short justifications of their decisions, we anticipate the data we collected can be used to further study this question in future work.

# References

Licato, J., Marji, Z., & Abraham, S. (2019). Scenarios and Recommendations for Ethical Interpretive AI. In *Proceedings of the AAAI 2019 Fall Symposium on Human-Centered AI*, Arlington, VA, 2019.

Licato, J. (2021). How Should AI Interpret Rules? A Defense of Minimally Defeasible Interpretive Argumentation. *arXiv e-prints*.

Marji, Z. and Licato, J. (2021). Aporia: The argumentation game. In *Proceedings of The Third Workshop on Argument Strength (ArgStrength 2021)*.

Walton, D., Macagno, F., and Sartor, G. (2021). *Statutory Interpretation: Pragmatics and Argumentation.* Cambridge University Press.

MacCormick, D. and Summers, R. (1991). *Interpreting Statutes: A Comparative Study.* Routledge.